\documentclass{article}
\usepackage{float} 
\usepackage{graphicx}
\usepackage{spconf,amsmath,graphicx,hyperref}
\usepackage{amssymb} 
\usepackage{tabularx} 
\usepackage{booktabs}
\usepackage{multirow}
\usepackage{afterpage}
\usepackage{stfloats}   

\title{MIRRORTALK: Forging Personalized Avatars Via Disentangled Style and Hierarchical Motion Control}
\name{Renjie Lu$^{1,2\ast}$, Xulong Zhang$^{1\ast}$, Xiaoyang Qu$^1$, Jianzong Wang$^{1 \dagger}$, Shangfei Wang$^{2 \dagger}$
\thanks{$^{\star}$ These authors contributed equally to this work.}%
\thanks{$^{\dagger}$ Corresponding author.}%
\thanks{Supported by Shenzhen-Hong Kong Joint Funding Project (Category A) under grant No. SGDX20240115103359001.}
}
\address{$^1$Ping An Technology (Shenzhen) Co., Ltd., Shenzhen, China \\
         $^2$University of Science and Technology of China, Hefei, China
         }
\begin{document}
\ninept
\maketitle
\begin{abstract}
Synthesizing personalized talking faces that uphold and highlight speaker’s unique style while maintaining lip-sync accuracy remains a significant challenge.\ A primary limitation of existing approaches is the intrinsic confounding of speaker-specific talking style and semantic content within facial motions, which prevents the faithful transfer of a speaker's unique persona to arbitrary speech.\ In this paper, we propose \textit{MirrorTalk}, a generative framework based on a conditional diffusion model, combined with the Semantically-Disentangled Style Encoder (SDSE) that can distill pure style representation from a brief reference video.\ To effectively utilize this representation, we then introduce a hierarchical modulation strategy within diffusion process.\ This mechanism guides the synthesis by dynamically balancing the contributions of audio and style features to distinct facial regions, ensuring both precise lip-sync and expressive full-face dynamics.\ Extensive experiments demonstrate that MirrorTalk achieves significant improvements against state-of-the-art methods in terms of lip-sync accuracy and personalization preservation.
\end{abstract}
\section{Introduction}
\label{sec:intro}

Audio-driven talking face generation is an inherently cross-modal task that aims to synthesize realistic talking videos by animating a target identity’s face according to arbitrary speech.\ Existing methods can be broadly categorized into two main paradigms.\ Some works \cite{Person-specific01, Person-specific02, Person-specific03} achieve high-fidelity talking face generation by performing person-specific training or fine-tuning, but they typically require extensive video data from the target speaker,\ limiting their scalability.\ Others \cite{uni02, wav2lip, universal, zhang2024emotalker, Liu2025ConsistTalkIC} focus on developing a universal model, which have recently achieved tremendous progress in lip synchronization, yet struggle to capture unique facial dynamics.\ This limitation stems from the tendency of models to learn a generalized mapping from audio to facial motion, which averages out the speaker-specific dynamics and results in homogeneous animations.

Generating expressive and realistic facial motions remains challenging, as it requires not only precise synchronization with the audio but also an accurate reflection of the target speaker's unique facial movements and expression variations.\ While several works \cite{emo01, mead} model talking style through a limited set of discrete emotion classes, this approach is often too coarse to capture the subtle and speaker-specific characteristics.\ Subsequent research \cite{styletalk, personatalk, styletalk++} explore the use of additional video clips as style references to guide the personalized generation of facial animations, which facilitates achieving expressive results.\ By deploying a style encoder to extract the speaker’s talking style from a reference video, these methods leverage it as a condition to modulate the generation.\ However, this paradigm suffers from a fundamental flaw: the entanglement of talking style and semantic content.\ The extracted style features is intrinsically confounded with the semantic content of the reference speech, creating an unstable and context-dependent representation.\ This leads to a conflict when the talking style is transferred to a new speech with different semantics, which resulting in degradation of lip-sync accuracy and unfaithful synthesis of facial motions.

To address these challenges and limitations, we propose Mirrortalk,\ a novel diffusion-based generative framework, which is capable of not only producing audio-synchronized lip movements, but also upholding speaker's unique talking style and facial dynamics.\ As its core, MirrorTalk first introduces a Semantically-Disentangled Style Encoder (SDSE).\ Trained via a unique two-stage, cross-modal strategy, the SDSE can distills a pure, content-agnostic style representation from a brief reference video, which is crucial for generating personlized and realistc facial motions.\ To effectively utilize this purified style representation, we then propose a spatial-temporal hierarchical modulation mechanism within a diffusion-based generating process to precisely guide the fusion of multi-modal conditions.\ This strategy enables the precise fusion of the style and audio features by dynamically balancing their contributions across different facial regions, which results in the synthesis of motion that is both precisely synchronized with the audio and faithful to the speaker's expressive style.\ Our main contributions are summarized as:
\begin{itemize}
\item[$\bullet$]We propose a novel two-stage disentangled representation learning framework, which effectively separates target's talking style from confounding semantic content with only a brief reference segment.
\item[$\bullet$]We introduce a spatial-temporal hierarchical modulation strategy for conditional diffusion models, enabling dynamic and region-aware balancing of audio and style features for expressive personalized motion synthesis.
\item[$\bullet$]Extensive experiments demonstrate that our method outperforms existing state-of-the-art approaches in lip-sync accuracy and personalization preservation.
\end{itemize}
\begin{figure*}[htb]
    \centering
    \includegraphics[width=0.95\textwidth]{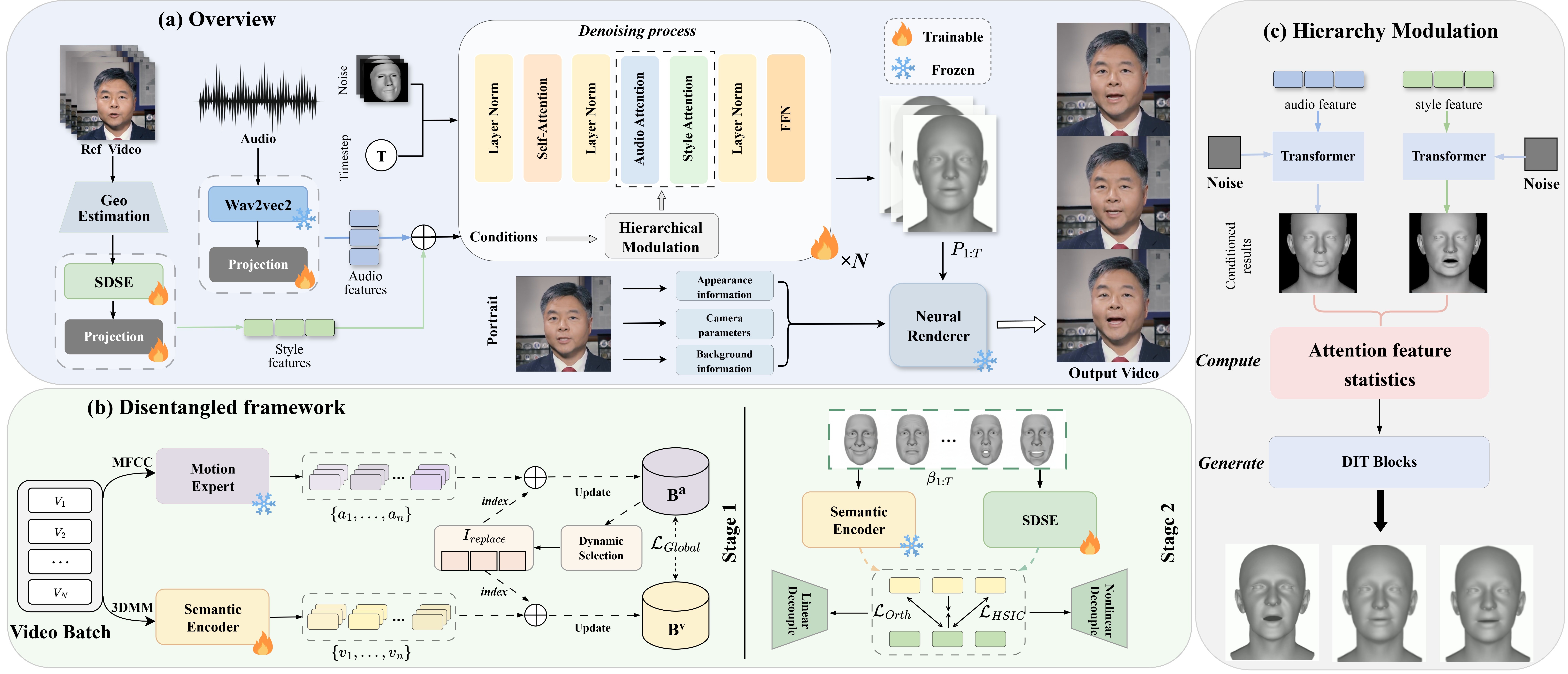}
    \vspace{-3mm}
    \caption{\textbf{Architecture of MirrorTalk.} We first introduce a two-stage training framework (b) to obtain the \textbf{Semantically-disentangled Style Encoder (SDSE)} for talking style prediction. In the main generation pipeline (a), audio and reference video inputs are encoded into compressed tokens as conditions for a diffusion transformer (DiT) model. During the denoising process, we employ a hierarchical modulation strategy (c) to dynamically balances the contributions of audio and style features for distinct facial regions at each timestep $T$. Finally,  a Neural Renderer \cite{pirenderer} utilizes the portrait image and generated motion sequence $P_{1:T}$ to synthesize the final video frames.}    \label{fig:overv}
\end{figure*}

\section{Method}
\label{sec:format}
An overview of our proposed method is illustrated in Fig. \ref{fig:overv}.\ Our framework firstly leverages a Semantically-Disentangled Style Encoder (SDSE) to obtain pure talking style representations from a reference video.\ We then utilize a spatial-temporal hierarchical modulation strategy to guide a diffusion-based generation, ensuring both lip-sync accuracy and hign realistic facial motions.\ These key contributions will be detailed in the following subsections.
\subsection{Facial geometry estimation}
\label{sec:problem}
For a given speaker video $V_{i}$, we can utilize a 3D morphable model FLAME \cite{flame} to obtain speaker's facial parameters $P_{1:T} = \{\{\alpha_{1}, \beta_{1}, \theta_{1}\}, \dots, \{\alpha_{T}, \beta_{T}, \theta_{T}\}\}$ of video frames $F_{1:T}$, where $\alpha_{t}\in\mathbb{R}^{100}$, $\beta_{t}\in\mathbb{R}^{50}$ and $\theta_{t}\in\mathbb{R}^{15}$ respectively denote shape, expression and pose.\ Specifically, to estimate these parameters from audio-visual inputs with high accuracy, we adopt SMIRK \cite{smirk} for accurate expression prediction, MICA \cite{mica} for identity-related shape estimation and 3DDFA \cite{3DDFA} for head pose reconstruction.\ Afterwards, following EmoTalk \cite{emotalk}, we apply a Savitzky–Golay smoothing filter to the estimated expression and pose parameters to improve motion smoothness.
\subsection{Disentangled Framework for Style Embedder}
\label{sec:SDSE}
The speaker’s facial motion $m_{i}$ is jointly determined by the semantic content $m_{i}^{semantic}$ that reflects the meaning of the content, and the individual’s speaking style $m_{i}^{style}$ that denotes the speaker-specfic motion patterns such as articulation habits and expression dynamics. We hypothesize that $m_{i} = m_{i}^{semantic} + m_{i}^{style}$, modeling their additive contribution to facial motion.
Previous methods \cite{styletalk, personatalk} for stylized facial animations often fail to explicitly disentangle semantic information from style representation. To address this, our approach begins with a style encoder backbone that extracts a comprehensive motion representation from a given reference video. Specifically, we use a transformer-based encoder that takes the sequential expression parameters $\beta_{1:T}$ as input. By modeling their temporal dependencies and employing a frame-level self-attention pooling layer, the encoder aggregates per-frame vectors $\mathbf{s}'_t$ into a overall style embedding $\mathbf{s}$ using the computed attention weight $e_{t}$:
\begin{equation}
\mathbf{s} = \sum_{t=1}^T \left( \frac{\exp(e_t)}{\sum_{k=1}^T \exp(e_k)} \right) \mathbf{s}'_t
\end{equation}
To isolate talking style representation from semantic content, we introduce a novel two-stage training strategy to disentangle them.

\textbf{Stage 1: Cross-Modal Supervision for Semantic Encoder Training.}
The primary objective of this stage is to train a semantic encoder to derive semantic representations from visual-only motion signals.\ To this end, a supervisory signal is required, for which we leverage a pre-trained Motion Expert.\ Its architecture is adapted from the powerful lip-sync discriminator of Wav2Lip \cite{wav2lip}.\ We takes pairs of audio MFCCs and corresponding lower-face image crops as input to retrain the model, endows the Motion Expert with a robust understanding of speech-related facial motion.\ This expert model serves to extract audio-semantic embeddings $a_i$ from the audio modality of $V_i$, which act as semantic target.\ The semantic encoder is then trained to produce visual-semantic embeddings $v_i$ from $\beta_{1:T}$ of $V_i$.\ To facilitate a robust alignment, we utilize memory banks $B^a$, $B^v$ and employ a redundancy-based update strategy to enhance model’s global perception.\ For each training batch, we first compute the redundancy $\rho_i$ for each embedding in $B^a$:
\begin{equation}
\rho_{i}=\frac{1}{N-1} \sum_{j=1, j \neq i}^{N} S_{i j}
\end{equation}
where $S_{ij} = \mathrm{sim}(a_i, a_j)$. We then identify the indices of the most redundant embeddings, denoted as $\mathbb{I}_{\text{replace}}$, and use them to update $B^a$ and $B^v$.\ The Semantic Encoder is ultimately trained by minimizing a global structural loss between the two spaces:
\begin{equation}
\mathcal{L}_{\text{global}} = \sum_{i, j \in \text{mem}} \left( \cos(v_i, v_j) - \cos(a_i, a_j) \right)^2
\end{equation}

\textbf{Stage 2: Semantic-Aware Style Disentanglement.}\ With the semantic encoder frozen, we train the SDSE to extract style representations that are disentangled from the semantic information.\ The training objective for the SDSE is a joint optimization of two losses.\ First, a decoupling loss is used to enforce independence between the style and semantic embeddings by combining an orthogonalization constraint and a nonlinear independence regularizer based on the Hilbert-Schmidt Independence Criterion (HSIC) \cite{hsic}.
\begin{equation}
\mathcal{L}_{\text{decouple}}
= \lambda_{\text{orth}} \left\| \tilde{\mathbf{s}}^\top \tilde{\mathbf{v}} \right\|_{F}^2
+ \lambda_{\text{hsic}} \, \mathrm{HSIC}(\tilde{\mathbf{s}}, \tilde{\mathbf{v}})
\end{equation}
where $\tilde{\mathbf{s}}$ and $\tilde{\mathbf{v}}$ denote the normalized style and visual-semantic embeddings.\ Second, to learn a style representation that is both highly discriminative between different speakers and consistent across various speech from the same speaker, we apply a triplet loss:
\begin{equation}
\mathcal{L}_{\text{triple}}
= \max\!\left(0,\; \delta + \left\| \mathbf{s}^{a} - \mathbf{s}^{p} \right\|_{2}^{2}
- \left\| \mathbf{s}^{a} - \mathbf{s}^{n} \right\|_{2}^{2}\right)
\end{equation}
where $\delta$ is the margin, and $\mathbf{s}^{a}$, $\mathbf{s}^{p}$, $\mathbf{s}^{n}$ are the anchor, positive, and negative style samples, respectively.\ The final SDSE model is obtained by minimizing the combined loss: $\mathcal{L}_{\text{total}} = \mathcal{L}_{\text{decouple}} + \mathcal{L}_{\text{triple}}$.

\begin{table*}[!t]
\centering
\caption{\textit{Quantitative comparisons with the state-of-the-arts on the CREMA-D and HDTF datasets. \textbf{Bold} indicates the best results , and the second value are \underline{underlined}.}}
\label{tab:baseline}
\resizebox{0.97\textwidth}{!}{
\small 
\setlength{\tabcolsep}{3.5pt} 
\begin{tabular}{@{} c *{12}{c} @{}}
\toprule
 & \multicolumn{6}{c}{CREMA-D} & \multicolumn{6}{c}{HDTF} \\ 
\cmidrule(lr){2-7} \cmidrule(l){8-13} 
Method & SSIM$\uparrow$ 
& FID$\downarrow$ 
& M-LMD$\downarrow$ 
& F-LMD$\downarrow$
& Sync$_{\text{conf}}\uparrow$
& StyleSim$\uparrow$ 
& SSIM$\uparrow$ 
& FID$\downarrow$ 
& M-LMD$\downarrow$ 
& F-LMD$\downarrow$
& Sync$_{\text{conf}}\uparrow$
& StyleSim$\uparrow$ \\ 
\midrule
Wav2Lip \cite{wav2lip} & 0.725 & 32.461 & \underline{3.025} & 3.476 & \textbf{4.384} & 0.826 & 0.618 & 38.744 & 4.121 & 4.040 & \underline{3.762} & 0.841\\
EAMM \cite{Ji2022EAMMOE} & 0.414 & 37.296 & 6.630 & 6.819 & 1.545 & 0.788 & 0.396 & 42.158 & 6.019 & 7.135 & 1.204 & 0.805\\
SadTalker \cite{sadtalker} & 0.762 & \textbf{15.135} & 4.143 & 2.804 & 2.676 & 0.851 & 0.664 & \textbf{20.514} & \underline{3.559} & 2.926 & 2.232 & 0.862\\
AniTalker \cite{anitalker} & 0.726 & \underline{16.141} & 5.742 & 4.052 & 1.926 & 0.730 & 0.593 & 25.259 & 6.413 & 4.547 & 2.763 & 0.724\\
Echomimic \cite{echomimic} & \underline{0.912} & 28.506 & 4.006 & \underline{2.612} & 3.461 & \underline{0.852} & \underline{0.879} & 31.243 & 3.681 & \underline{2.851} & 2.689 & \underline{0.866}\\
V-Express \cite{vexpress} & 0.708 & 18.074 & 4.906 & 4.868 & 2.130 & 0.834 & 0.651 & 24.061 & 5.706 & 5.001 & 1.593 & 0.845\\
\midrule
Ours & \textbf{0.917} & 16.293 & \textbf{2.771} & \textbf{1.824} & \underline{4.106} & \textbf{0.937} & \textbf{0.890} & \underline{21.682} & \textbf{2.481} & \textbf{2.122} & \textbf{3.811} & \textbf{0.958}\\
Ground Truth & 1.000 & 0.000 & 0.000 & 0.000 & 4.531 & 0.942 & 1.000 & 0.000 & 0.000 & 0.000 & 3.962 & 0.969\\
\bottomrule
\end{tabular}
}
\end{table*}

\subsection{Hierarchical Conditioning for Motion Synthesis}
\label{sec:diffusion}
We employ a diffusion transformer model \cite{dit} to generate the motion sequence.\ The model is trained to denoise a noisy sample $x_t$ back to the original data $x_0$ conditioned on features $c$, which is achieved by training a network $\epsilon_\theta$ to predict the added noise $\epsilon$ at each timestep $t$, optimized via the following objective:
\begin{equation}
\mathcal{L}_{\text{denoising}} = \mathbb{E}{ x_0, \epsilon, c, t} \left\|\epsilon - \epsilon\theta(x_t, c, t) \right\|^2
\end{equation}

Previous works used to directly inject $c$ into diffusion model through cross attention. However, the motion patterns of the upper and lower face exhibit distinct characteristics: the upper face is predominantly influenced by speaking style $c_s$, whereas the lower face is more strongly correlated with audio feature $c_a$. To account for this differential behavior, we propose a spatial–temporal hierarchical strategy to dynamically modulate audio and style features.\ For each of the two facial regions $r$—the upper face $r_u$ and the lower face $r_l$—at every denoising time step $t$, we first compute the cosine similarity between the outputs of the audio-conditioned cross-attention $Z_a(r,t)$ and style-conditioned cross-attention $Z_s(r,t)$ with respect to the combined feature map $Z(r,t)$:
\begin{equation}
\begin{aligned}
&P_{a}(r, t) = \cos\left(Z_{a}(r, t), Z(r,t)\right)\\
&P_{s}(r, t) = \cos\left(Z_{s}(r, t), Z(r,t)\right)
\end{aligned}
\end{equation}
While audio features often exhibit stronger influence ($P_a > P_s$), the magnitude of this dominance varies significantly across timesteps. A static correction would fail to capture this nuance, potentially over-suppressing audio during crucial early structural formation or under-amplifying style during late-stage refinement. To address this, we introduce $D(r, t)$ which measures the relative dominance of audio over style, that adapts at region $r$ and timestep $t$:
\begin{equation}
\begin{aligned}
D(r, t) = \sigma \left( P_a(r, t) - P_s(r, t) \right)
\end{aligned}
\end{equation}
where $\sigma$ is the sigmoid function.\ This factor then re-weights the contributions of $Z_a$ and $Z_s$ based on the spatial prior of each facial region,\ producing the modulated feature map $Z'(r,t)$ as follows:
\begin{equation}
Z'(r,t) = 
\begin{cases} 
Z_s(r, t) / D(r, t) + Z_a(r, t), & \text{if } r = r_u \\
Z_a(r, t) * D(r, t) + Z_s(r, t), & \text{if } r = r_l
\end{cases}
\end{equation}

\begin{table}[!t]
\centering
\caption{\textsc{Ablation studies on our method.\ \textbf{Bold} means the best.}}
\label{tab:ablation}
\resizebox{0.485\textwidth}{!}{
\begin{tabular}{lcccc}
\toprule
\multicolumn{1}{c}{Ablation} & M-LMD$\downarrow$ & F-LMD$\downarrow$ & Sync$_{\text{conf}}\uparrow$ & StyleSim$\uparrow$ \\
\midrule
\multicolumn{1}{c}{w/o Memory Bank} & 3.074 & 2.426 & 3.473 & 0.869 \\
\multicolumn{1}{c}{w/o Dis-Module} & 3.687 & 2.581 & 2.805 & 0.837 \\
\multicolumn{1}{c}{w/o $\mathcal{L}_{\text{triple}}$} & 2.933 & 2.734 & 3.724 & 0.901 \\
\multicolumn{1}{c}{w/o H-Scales} & 3.281 & 2.401 & 3.059 & 0.911 \\
\midrule
\multicolumn{1}{c}{\textbf{Ours(Full Model)}} & \textbf{2.503} & \textbf{2.265} & \textbf{3.843} & \textbf{0.938} \\
\bottomrule
\end{tabular}
}
\end{table}

\section{EXPERIMENT}
\label{sec:experiment}

\subsection{Datasets}
\label{sec:setup}
We leverage a composite dataset consisting of VoxCeleb2 \cite{voxceleb2}, HDTF \cite{HDTF} and CREMA-D \cite{crema}.\ Specifically, VoxCeleb2 is an extensive audio-visual library featuring 1 million+ utterances from 6,112 YouTube figures. HDTF is a high-resolution dataset that contains 16 hours of videos. CREMA-D is an emotional talking-face dataset involving 91 identities.\ Preprocessing involves resampling all videos to 25 fps and then cropped to the size of 512$\times$512.

\subsection{Experiment setup}
\textbf{Evaluation Metrics.}\ Consistent with previous works \cite{guanmetric, styletalk++}, we use Structured Similarity (\textbf{SSIM}) \cite{ssim} and Frechet Inception Distance (\textbf{FID}) to access the visual fidelity.\ For account of lip-sync accuracy, we utilize the Landmark Distance around the mouth (\textbf{M-LMD}) \cite{mlmd} and the confidence score of SyncNet ($\mathbf{Sync}_{\mathbf{conf}}$) \cite{sync}. Furthermore, we employ two metrics - the Landmark Distance on the whole face (\textbf{F-LMD}) and Speaking Style Similarity (\textbf{StyleSim}) \cite{personatalk} - to measure preservation of speaker‘s persona.

\textbf{Baselines.}\ Our comparative analysis includes several state-of-the-art person-generic methods of different types, including Wav2Lip \cite{wav2lip}, EAMM \cite{Ji2022EAMMOE}, AniTalker \cite{anitalker}, SadTalker \cite{sadtalker}, Echomimic \cite{echomimic}, V-Express \cite{vexpress}.\ Wav2lip utilizes a pretrained lip-expert to train the model for improving the audio-visual synchronization.\ AniTalker employs self-supervised learning to disentangle and generate identity-agnostic facial motion.\ SadTalker innovates by separately modeling audio-to-expression and audio-to-head pose to drive a 3D-aware neural renderer.\ Echomimic combines audio inputs with reference landmarks to produce naturalistic results. These baselines are selected to provide a comprehensive evaluation across various aspects of talking-face generation.
\begin{figure*}[htb] 
    \centering
    \includegraphics[width=0.92\textwidth]{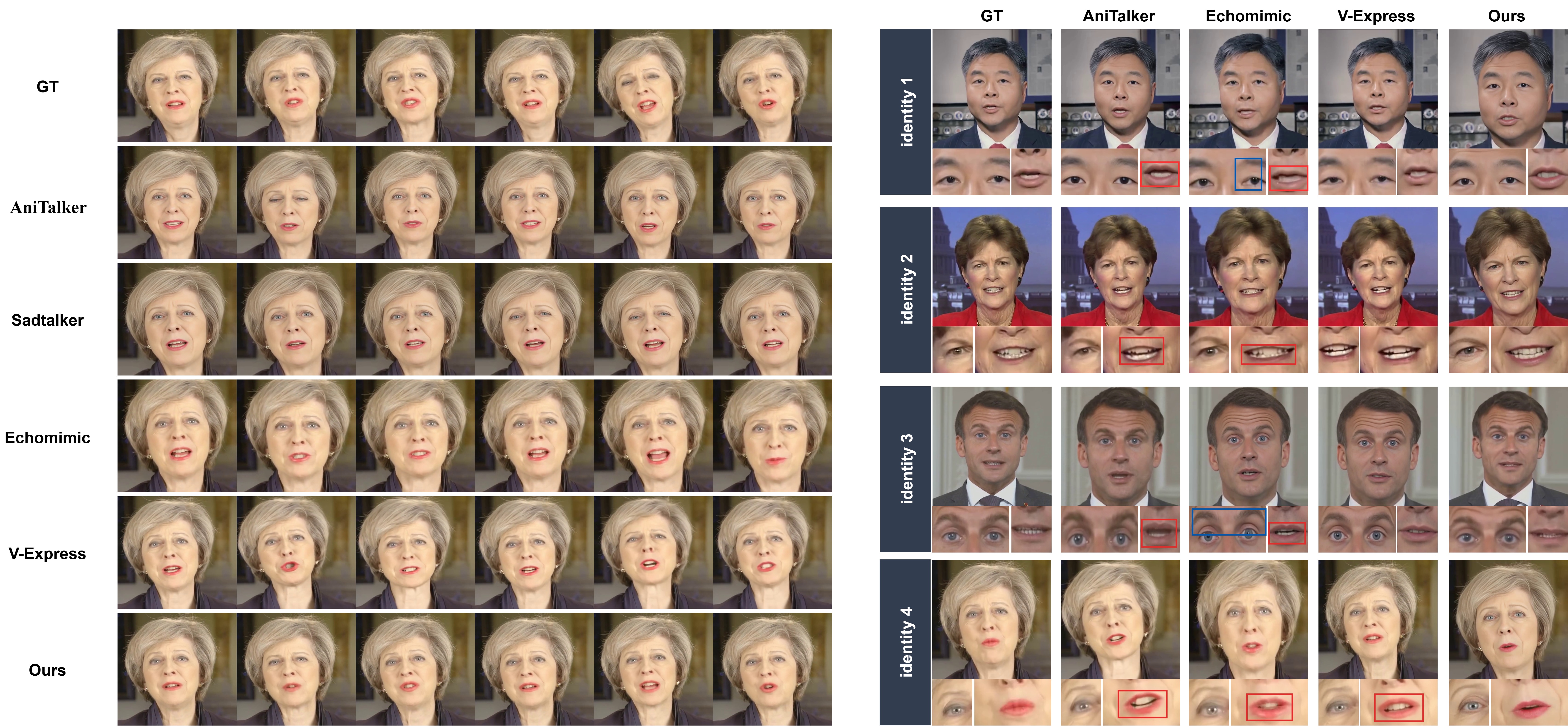}
    \vspace{-3mm}
    \caption{Qualitative comparsion with AniTalker \cite{anitalker}, SadTalker \cite{sadtalker}, Echomimic \cite{echomimic} and V-Express \cite{vexpress}. Red and blue boxes highlight incorrect lip movements and facial expressions in the synthesized image respectively. Our methods not only generates accurate lip movements, but also preserves speaker's talking style and facial dynamics.}
    \label{fig:lip_sync_comparison}
\end{figure*}

\subsection{Evaluation}
\textbf{Quantitative Evaluation.}\ As shown in Table \ref{tab:baseline}, our method obtain consistent improvements against other baselines on most metrics across both datasets. Regarding visual qualities, our approach achieves superior and competitive results compared to other methods.\ As for the lip-sync accuracy, we gets much better comparable performance in the M-LMD, which can be largely attributed to hierarchical scales that biases the lower face to audio cues for finer lip motion. Of particular note, in terms of talking style and facial dynamic preservation, our method is far ahead of other approachs.\ This superiority stems from the disentangled framework for style embedder, which yields a pure style representation, free from confounding semantic information.\ These indicates the progressiveness of our disentangled-framework for style embedder and hierarchical scales in capturing person-specific talking style and generating realist facial expressions.

\vspace{-1pt}
\textbf{Quanlitative Evaluation.}\ To more intuitively evaluate viusal effects, we display a comparison between our method and others in Fig.\ \ref{fig:lip_sync_comparison}.\ It can be observed that MirrorTalk demonstrates more accurate facial movements.\ Compared to AniTalker \cite{anitalker}, which generates rigid facial motions that lack a personal style, our methods also excels in lip shape synchronization. Against Sadtalker \cite{sadtalker} and Echomimic \cite{echomimic}, MirrorTalk can generate more expressive facial animations due to the hierarchical modulation strategy, particularly in the upper-face regions like the eyebrows and eyes.\ Compared to V-Express \cite{vexpress}, our methods achieves better preservation of target's speaking style.\ Overall, MirrorTalk achieves a superior balance of lip-sync accuracy and personalized expression within these methods.

\vspace{-1pt}
\textbf{Ablation Study.}\ We conduct an ablation study in Table \ref{tab:ablation}, to examine the contributions of different components in our model and select four core metrics for evaluation.\ Specifically, we first remove the memory bank in the training stage of semantic encoder. We observed that both M-LMD and F-LMD increase noticeably, and StyleSim also drops, which indicates that the absence of informative memory reduces the encoder’s ability to align semantic features.\ Besides, eliminating the disentanglement module causes the most severe performance degradation across all metrics, demonstrating the necessity of explicitly separating style from semantics.\ Moreover, removing the triplet loss $\mathcal{L}_{\text{triple}}$ results in a significant drop in StyleSim and the largest increase in F-LMD, underscoring its crucial role in learning a speaker-discriminative style representation.\ Finally, when the hierarchical scale modulation is removed, $\mathbf{Sync}_{\mathbf{conf}}$ decreases significantly along with a clear rise in landmark distances, proving that this strategy is critical for generating natural and realistic facial motions.critical for generating natural and realistic facial motions.
\section{Conclusion}
\label{sec:typestyle}
In this paper, we propose MirrorTalk, a novel diffusion-based framework for generating personalized facial animations.\ By disentangling style from content via our two-stage training framework, and fusing it with audio using a hierarchical strategy, our approach faithfully preserves a speaker’s unique persona.


\bibliographystyle{IEEEbib}
\bibliography{refs}

\end{document}